\documentclass[a4paper]{article}

\usepackage{INTERSPEECH2018}
\usepackage{flushend}

\title{Device-directed Utterance Detection}
\name{Sri Harish Mallidi, Roland Maas, Kyle Goehner,  \\ Ariya Rastrow, Spyros  Matsoukas, Bj\"orn Hoffmeister}
\address{Amazon, USA}
\email{\{mallidih,rmaas,kgoehner,arastrow,matsouka,bjornh\}@amazon.com}

\begin{document}

\maketitle
\begin{abstract}
In this work, we propose a classifier for distinguishing device-directed queries from background speech in the context of interactions with voice assistants.
Applications include rejection of false wake-ups or unintended interactions as well as enabling wake-word free follow-up queries.
Consider the example interaction: {\it ``Computer, play music'', ``Computer, reduce the volume''}.
In this interaction, the user needs to repeat the wake-word ({\it Computer}) for the second query. To allow for more natural interactions, the device could immediately re-enter listening state after the first query (without wake-word repetition) and accept or reject a potential follow-up as device-directed or background speech.
The proposed model consists of two long short-term memory (LSTM) neural networks trained on 
acoustic features and automatic speech recognition (ASR) 1-best hypotheses, respectively. 
A feed-forward deep neural network (DNN) is then trained to combine the acoustic and 1-best embeddings, derived from the LSTMs, with features from the ASR decoder. 
Experimental results show that ASR decoder, acoustic embeddings, and 1-best embeddings yield an equal-error-rate (EER) of $9.3~\%$, $10.9~\%$ and $20.1~\%$, respectively. 
Combination of the features resulted in a $44~\%$ relative improvement and a final EER of $5.2~\%$.
\end{abstract}
\noindent\textbf{Index Terms}: speech recognition, human-computer interaction, computational paralinguistics

\section{Introduction}
\label{sec:intro} 
The popularity of voice controlled far-field devices (e.g. Amazon Echo, Google Home) is on rise. 
These devices are often used in challenging acoustic environments. 
One such example is a living room scenario, where the device may capture speech from several speakers (not just the device user's speech) and multi-media speech (speech from TV or music player). 
In these situations, it is crucial for the device to act only on the intended (referred to as device-directed) speech and ignore un-intended (referred to as nondevice-directed) speech. 
In order to make the voice controlled device robust to nondevice-directed speech, a reliable device-directed vs nondevice-directed classifier is required, which we will investigate in this work. 

Past research on device-directed speech detection relied on acoustic features in addition to features from an ASR decoder \cite{learning-when-to-listen-detecting-system-addressed-speech-in-human-human-computer-dialog, reich2011real, yamagata2009system, lee2013using, wang2013understanding}.
Acoustic features used in these works are primarily related to prosodic structure of input speech. 
Energy and pitch trajectories, speaking rate, and duration information are computed and several statistics of these features are used for device-directed speech detection \cite{learning-when-to-listen-detecting-system-addressed-speech-in-human-human-computer-dialog}. 
Non-traditional acoustic features such as multi-scale Gabor wavelets have also been explored in \cite{yamagata2009system}. 
Features from non-acoustic sources like ASR confidence scores, N-grams have been used in \cite{learning-when-to-listen-detecting-system-addressed-speech-in-human-human-computer-dialog, yamagata2009system, dowding2006you}. 
A variety of classifiers are then used to model the extracted features, and their decisions are combined during inference. 

In this work, we use three sources of information: (i) acoustics, (ii) ASR decoder, and (iii) ASR 1-best hypothesis. 
Two LSTM models are trained on acoustics and 1-best word/character sequences, respectively, to obtain fixed length embeddings. 
A single feature vector per utterance is then constructed by concatenating these embeddings with features from ASR decoder. 
A fully connected neural network is trained on utterance feature. 
%
%

The paper is organized as follows:
Section 2 provides an overview of the proposed device-directed model. 
In this section, we also discuss the components in the model. 
Experimental analyses and results of the proposed model are discussed in Section 3. 
This section also provides details of the dataset used to train the models. 
Section 4 presents conclusions of the paper. 

\section{System architecture}

\begin{figure}[t!]
	\centering
	\includegraphics[width=\columnwidth]{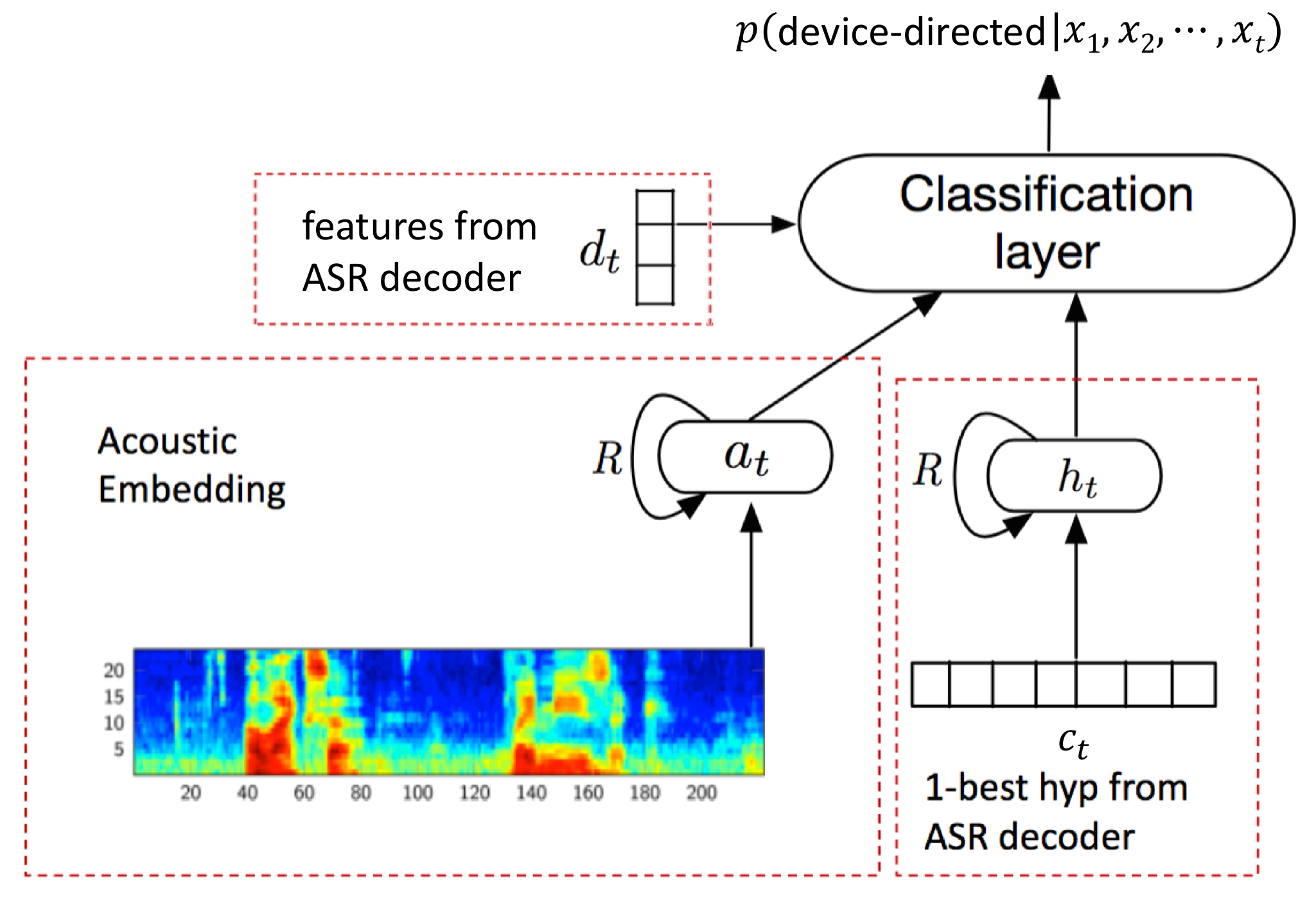}
	\caption{Proposed device-directed model, based on the combination of three features types: acoustic embedding, 1-best hypothesis embedding, and decoder features. }
	\label{fig:features}
\end{figure}

Figure~\ref{fig:features} illustrates the model we use for the device-directed detection. 
This architecture has been proposed for end-pointing task in \cite{EP2.0}.   
The model consists of 4 main components (i) acoustic embedding using LSTM, (ii) ASR decoder features, (iii) character embedding using LSTM, and (iv) classification layer. 
 
\subsection{Acoustic embedding}
The task here is to assign a single label (device-directed or nondevice-directed) to the given utterance. 
A fixed length vector of the utterance is obtained as follows: 
Short-term acoustic features, i.e. log filter-bank energies (LFBEs), are obtained using analysis window length of 25 milliseconds and 10 milliseconds shift. 
An LSTM model is trained on LFBEs to predict frame-level device-directed targets. 
The frame-level targets are obtained by repeating the utterance label. 
The parameters of the LSTM are optimized by minimizing the cross-entropy loss, using stochastic gradient descent \cite{lecun1998efficient, Strom2015ScalableDD}. 
We use the pre-softmax output of the last frame of input utterance as its representation, since it encodes all the information in the utterance (due to recurrence). 
This 2-dimensional vector is referred to as {\it acoustic embedding} ($\mathbf a$) of the utterance. 
  

%

\begin{table}[b!]
	\renewcommand{\arraystretch}{1.5}
	\caption{\label{tab:acousticLSTM_modelArch_analysis} Performance of acoustic LSTM model with respect to number of layers and number of cells in each layer.}
	\centerline{
		\begin{tabular}{ c @{\hskip10pt} c @{\hskip6pt} c }
			\hline
			acoustic LSTM & 
			EER(\%) & 
			\# parameters \\
			(\#layers$\times$\#cell size) & ~ & ~ \\
			\hline \hline
			$3\times768$ & $13.1$   & $12 M$ \\
			$4\times768$ & $10.9$   & $16 M$ \\
			$5\times768$  & $11.0$  &  $21 M$\\
			$6\times768$  & $11.1$  &  $26 M$\\
			$7\times768$  & $11.4$   & $31 M$\\
			\hline
		\end{tabular}
	}
\end{table}

\subsection{ASR decoder features}
Along with LSTM embeddings, we also use features obtained from ASR decoder. These are described below:

  In ASR, trellis can be used to efficiently estimate HMM parameters and infer the most likely state sequence. 
  Trellis structure can be used to effectively compute forward probabilities.   
  Entropy of the forward probability distribution is computed at every frame, and these are averaged. 
  A trellis with high entropy indicate that the probability mass is spread over alternate hypotheses, and ASR being less confident about its best hypothesis. 
  This can be due to language model mismatch or acoustic mismatch, which typically indicates nondevice-directed speech.

Along with trellis entropy, we extract Viterbi costs \cite{gales2008application}. 
  These features indicate how well the input acoustics and vocabulary match with the acoustic and language models. 
  Higher cost typically indicate greater mismatch between the model and given data. 

A confusion network \cite{mangu2000finding} is a simple linear graph, which is used as an alternative representation of the most likely hypotheses of the decoder lattice. 
The arcs in the confusion network correspond to words. 
Along with the word ids on each arc, confusion network also contains posterior estimates of each word.
ASR confidence of 1-best ASR hypothesis is obtained by taking a geometric product of all the posterior probabilities of words in the 1-best hypothesis \cite{evermann2000posterior}. 
Along with ASR confidence, we compute the average number of arcs from each node in the confusion network.
This relates to the number of competing hypotheses in the confusion network. 
Large number of competing hypotheses could indicate that the ASR system is being less confident about the 1-best hypothesis.
%
In total we use 18 features from the decoder. We used our in-house ASR system based on \cite{parthasarathi2015fmllr, garimella2015robust} to extract these features, referred to as {\it decoder features} (${\mathbf d}$). 


\begin{table}[]
\centering
\caption{Examples of 1-best hypotheses of device-directed and nondevice-directed utterances.}
\label{tab:d_vs_nd_asr_1best}
\begin{tabular}{|l|}
\hline
{\bf Device-directed speech} \\ \hline \hline 
what's the weather like in las vegas                                      \\ \hline
play popular music                                                        \\ \hline
mark the first item done                                                  \\ \hline
what is scratch programming                                               \\ \hline \hline
{\bf Nondevice-directed speech} \\ \hline \hline 
or if they want she can just queue for better                             \\ \hline
well that's how we had all the training                                   \\ \hline
talk to alexa but we're talking to danny right now                        \\ \hline
live together like months ago and they may still be \\ boring                \\ \hline
\end{tabular}
\end{table}

\begin{table}[b!]
	\renewcommand{\arraystretch}{1.5}
	\caption{\label{tab:glove_dims} Performance of char LSTM model with respect to the size of character embedding.}
	\centerline{
		\begin{tabular}{ c @{\hskip10pt} c  }
			\hline
			Embedding dimension & 
			EER(\%) \\
			\hline \hline
			$~~50$ & $21.0$    \\
			$100$   & $20.6$  \\
			$200$   & $20.1$    \\
			$300$   & $21.3$    \\
			\hline
		\end{tabular}
	}
\end{table}


\begin{table}[b!]
	\renewcommand{\arraystretch}{1.5}
	\caption{\label{tab:errorRates} Device-directed performance using various features.}
	\centerline{
		\begin{tabular}{ l  c @{\hskip6pt}}
			\hline
			features & 
			EER(\%) \\
			\hline \hline
			decoder features (${\mathbf d}$)         & $~~9.3$   \\
			acoustic embedding  (${\mathbf a}$)   & $10.9$   \\
			char embedding  (${\mathbf c}$)          & $20.1$  \\
			\hline
			$[\mathbf {a, d}]$          & $~~6.5$   \\
			$[\mathbf {c, d}]$          & $~~6.9$   \\
			$[\mathbf {a, c}]$          & $~~8.6$  \\
			\hline 
			$[\mathbf {a, c, d}]$      & $~~5.2$   \\
			\hline
		\end{tabular}
	}
\end{table}

\subsection{Character embedding}
Similar to {\it acoustic embedding}, we extract a fixed length representation from ASR 1-best hypotheses. 
Character sequence of a 1-best hypothesis is converted into vector sequence using pre-trained embedding vectors. 
We use GloVe embeddings \cite{pennington2014glove} for this purpose. 
An LSTM is trained on the vector sequence to predict frame-level device-directed decisions. 
Note that frame here refers to a character in the 1-best hypothesis. 
Once the network is trained, the network output of the last character is used as representation of 1-best hypothesis. 
This is referred to as {\it char embedding} ($\mathbf c$) of the input utterance.  

Once the acoustic and char LSTMs \cite{hochreiter1997long} are trained, the 2 embeddings are extracted and concatenated with decoder features to form a 22-dimensional utterance vector (${\mathbf f} = [{\mathbf a},~ {\mathbf c},~ {\mathbf d}] $). 
This vector is used as input to train a fully connected network (classification layer in figure~\ref{fig:features}). 

\section{Experiments}
We use real recordings of natural human interactions with voice-controlled far-field devices for training and testing the models. 
The training dataset consists of 250 hours of audio data comprised of 350k utterances. 
Of these, 200k and 150k are device-directed and nondevice-directed examples, respectively. 
The test data consists of 50k utterances (30 hours of audio data), with 38k device-directed and 12k nondevice-directed utterances.  
The classification performance is evaluated in terms of equal-error-rate (EER \%). 
EER correspond to the point on detection-error-tradeoff (DET) curve where false positive rate is equal to false negative rate. 
We also report DET curve to asses whether the improvement is consistent across several operating points. \\

\subsection{LSTM architecture}
{\bf Acoustic LSTM}: We used 64 dimensional LFBEs to train an acoustic LSTM. 
Number of layers in the network are varied to find the optimal model architecture. 
%
Table~\ref{tab:acousticLSTM_modelArch_analysis} shows the EERs of several acoustic LSTMs. 
It can be inferred from the table that adding more layers to the model improve the performance. 
Lowest EER is obtained by the model with 4 layers ($16M$). 
Adding more layers does not improve the performance. 
We hypothesize that this might be due to not enough training data.  \\

\begin{figure}[t!]
	\centering
	\includegraphics[width=0.9\columnwidth]{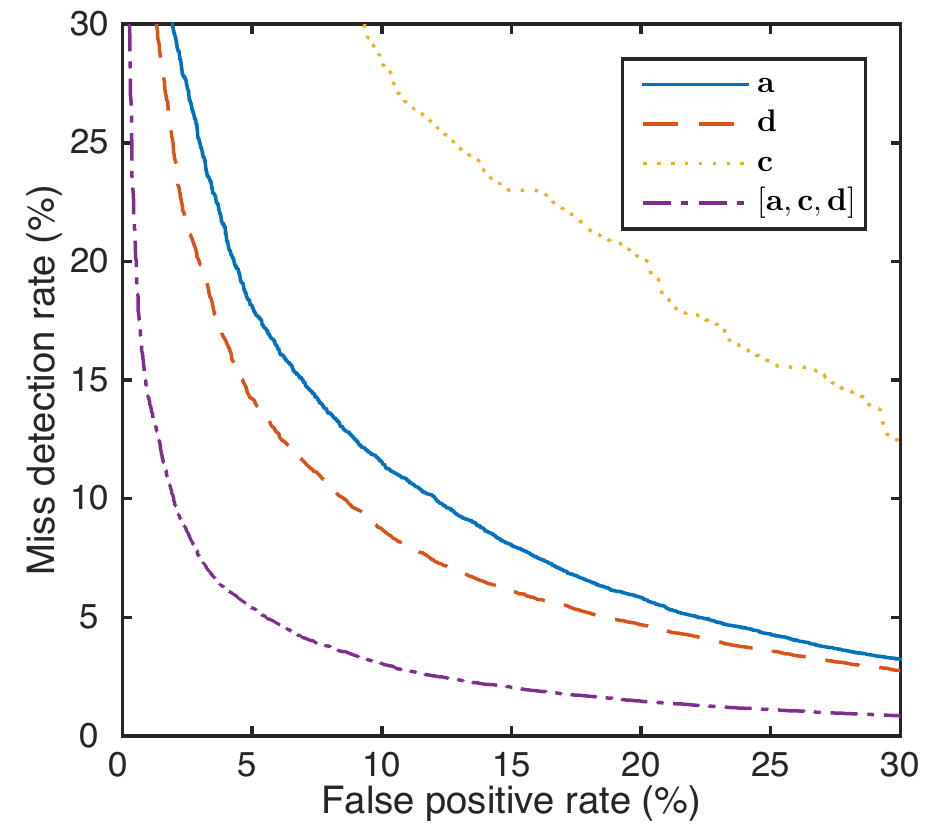}
	\caption{Detection error tradeoff (DET) curves of {\it acoustic embedding} (${\mathbf a}$), {\it decoder features} (${\mathbf d}$), {\it char embedding} (${\mathbf c}$), and {\it combination model} ($\mathbf{[a, c, d]}$).}
	\label{fig:final_det_curve}
\end{figure}

\noindent
{\bf Char LSTM}: A deeper analysis of device-directed and nondevice-directed utterances indicated that, device-directed speech is structured, containing short phrases and similar words. 
In comparison, nondevice-directed speech is more spontaneous and less grammatical. 
This is reflected even in 1-best hypothesis of ASR.
Table~\ref{tab:d_vs_nd_asr_1best} illustrates a few1-best hypotheses. 
We can infer from the table that 1-best hypotheses of device-directed speech is markedly different from nondevice-directed speech. 
Inspired by this, we use embedding of 1-best hypothesis. 
Table~\ref{tab:glove_dims} shows the EERs of char LSTM models as a function of input embedding size. 
It can be inferred from the table that a 200 dimensional embedding is optimal for this task. 

\subsection{Results}
Comparison between tables~\ref{tab:acousticLSTM_modelArch_analysis} and~\ref{tab:glove_dims} shows that the acoustic LSTM is performing better than char LSTM. 
Lower accuracy of char LSTM might due to a smaller amount of training data.  
The acoustic LSTM is trained on $91$ million frames where as char LSTM is trained on $8$ million frames.

Table~\ref{tab:errorRates} shows the performance of the acoustic LSTM, char LSTM, decoder features, and their combinations. 
It can observed that decoder features are best performing ($9.1~\%$ EER) followed by acoustic LSTM ($11~\%$ EER) and char LSTM ($20~\%$ EER). 
It can also be inferred from the table that the features are complimentary in nature since the combination of features never degrades over the single features. 
The final performance of the proposed model is $5.2 \%$ EER.  
Figure~\ref{fig:final_det_curve} shows the DET graphs of individual features and the final combination. 
It can be inferred that the combination performs better than individual features in all operating points. 

%

\section{Conclusions}
In this work, we explored acoustic features and features from ASR system for the task of device-directed utterance detection. 
An acoustic embedding vector is extracted by training an LSTM network on LFBEs. 
From ASR system, we extract several decoder features (trellis entropy, Viterbi costs, etc.). 
Also, we model 1-best hypothesis by first transforming the character sequence into vector sequence (using GloVe embeddings). 
An LSTM is trained on the resulting vector sequence to obtain a char embedding. 
Experimental results indicate that acoustic embedding, char embedding and decoder features are useful for device-directed task. 
The combination of all 3 features resulted in an $44~\%$ relative reduction EER over best individual feature ($9.3~\%$ to $5.2~\%$ EER), indicating complementary nature of the features.

\bibliographystyle{IEEEtran}

\bibliography{combined}


\end{document}